\documentclass[a4paper,twoside]{article}

\usepackage{epsfig}
\usepackage{subcaption}
\usepackage{calc}
\usepackage{amssymb}
\usepackage{amstext}
\usepackage{amsmath}
\usepackage{amsthm}
\usepackage{multicol}
\usepackage{pslatex}
\usepackage{apalike}
\usepackage{algorithm2e}
\usepackage[bottom]{footmisc}

\usepackage{cite}
\usepackage{amsmath,amssymb,amsfonts}
\usepackage{algorithmic}
\usepackage{graphicx}
\usepackage{textcomp}
\usepackage{xcolor}
\usepackage{cite}
\usepackage{svg}

\usepackage{url}
\usepackage{multirow}

\usepackage{SCITEPRESS}     

\begin{document}

\title{Privacy Risk Predictions Based on Fundamental Understanding of Personal Data and an Evolving Threat Landscape}

\author{\authorname{Haoran Niu\sup{}\orcidAuthor{0000-0003-4596-6127}, K. Suzanne Barber\sup{}\orcidAuthor{0000-0003-2906-6583}}
\affiliation{\sup{}The University of Texas at Austin}
\email{haoranniu@utexas.edu, sbarber@identity.utexas.edu}
}



\abstract{It is difficult for individuals and organizations to protect personal information without a fundamental understanding of relative privacy risks. By analyzing over 5{,}000 empirical identity theft and fraud cases, this research identifies which types of personal data are exposed, how frequently such exposures occur, and what the consequences of those exposures are. We construct an Identity Ecosystem graph---a foundational, graph-based model in which nodes represent personally identifiable information (PII) attributes and edges represent empirical disclosure relationships between them (e.g., one PII attribute is exposed due to the exposure of another). Leveraging this graph structure, we develop a privacy risk prediction framework that uses graph theory and graph neural networks to estimate the likelihood of further disclosures when certain PII attributes are compromised. The results show that our approach effectively addresses the core question: Can the disclosure of a given identity attribute possibly lead to the disclosure of another attribute? The code for the privacy risk prediction framework is available at: \url{https://github.com/niu-haoran/Privacy-Risk-Predictions-and-UTCID-Identity-Ecosystem.git}.}

\onecolumn \maketitle \normalsize \setcounter{footnote}{0} \vfill

\section{\uppercase{Introduction}}
\label{introduction_sec}

Different individuals and organizations have different sets of personally identifiable information (PII), and therefore hold distinct perspectives on which PII attributes are more vulnerable, more valuable, and in greater need of protection. An individual’s PII includes personal data in four categories---What You Know (e.g., name, address, phone number, mother’s maiden name), What You Have (e.g., driver’s license, Social Security card, employee ID, passport), What You Are (e.g., fingerprint, voice, facial image), and What You Do (e.g., patterns of life such as websites visited, GPS location history, phone logs) \cite{zaeem2016predicting}. 

Protecting PII data can be costly and time-consuming. Previous research has uncovered various strategies to reduce the risks of unintended data disclosure \cite{karr2014using}, including statistical disclosure limitation (SDL) techniques commonly used by national statistical agencies before releasing public-use data sets. Meanwhile, research on data self-destruction focuses on protecting data privacy for users who choose cloud services \cite{zeng2010safevanish, geambasu2009vanish}. Among the many existing privacy protection methodologies, this paper focuses on the first step of privacy protection: determining which set of data to protect. Protecting the most valuable and risky (i.e., likely to be exposed) set of PII promises to be a more effective and efficient method of protecting a person’s personal, sensitive data. This is because individuals and institutions usually have limited time, energy, and financial resources allocated for privacy protection. 

This research is based on the premise that if individuals and organizations have a more fundamental understanding and a more accurate evaluation of privacy risks resulting from disclosing PII, they will be in a better position to protect that information. Additionally, better risk analysis of personal data sharing will inform a wide range of information security and privacy applications. 

Accordingly, this research determines which data require the most protection by evaluating the consequences of exposing specific PII attributes. Specifically, it analyzes the privacy risks incurred when an individual or an organization shares or loses a given PII attribute. We primarily seek to answer the question: \textbf{Could the disclosure of a given PII attribute, such as date of birth, lead to the disclosure of another attribute, such as an ATM PIN}? PII attributes have associated risk values \cite{zaeem2016predicting, zaiss2019identity}. By evaluating the risk scores of the potentially disclosed PII attributes, we provide a quantified prediction of privacy risks based on empirical data on disclosures. 

To provide quantified predictions of privacy risks, we leverage graph theory. Many well-studied networks, such as transportation networks and social networks, are analyzed using graphs \cite{nippani2023graph, ediger2010massive}. Similar to these systems, meaningful connections and relationships exist among PII attributes. This research represents each PII attribute as a node in the graph. The directional edges capture the disclosure/exposure relationship between two PII attributes---specifically, the probability that the disclosure of one PII attribute (e.g., date of birth) could lead to the exposure of another PII attribute (e.g., address). We refer to this graph of PII nodes with directed disclosure or exposure relationships as the University of Texas Center for Identity (UTCID) Identity Ecosystem graph.

After constructing UTCID Identity Ecosystem graphs, we create and train three different link prediction models. We also develop a risk score calculation model. Together, the UTCID Identity Ecosystem graphs, the link prediction algorithms, and the risk score calculation model form a comprehensive risk prediction framework. 

Suppose an individual has lost PII attributes $A_1$ and $B_2$ and wishes to determine which other PII attributes become highly risky as a result. The pipeline of the risk prediction framework is outlined below:

\begin{itemize}
  \item The individual provides a risk score threshold (e.g., on a $[0, 100]$ scale) to indicate the level of risk of concern. The default risk score threshold is $0$, meaning that all potentially disclosed attributes are considered.
  \item The individual specifies the PII attributes $A_1$ and $B_2$ that were stolen or lost. 
  \item The PII attribute set is defined as all nodes in the UTCID Identity Ecosystem graph associated with the individual, excluding $A_1$ and $B_2$. Alternatively, the individual may specify separate PII attribute sets of interest for $A_1$ and $B_2$. 
  \item The risk prediction model---composed of a link prediction module and a risk score calculation module---takes $A_1$, $B_2$, the risk score threshold, and the PII attribute set as input, and outputs the PII attributes that may be disclosed and have risk scores exceeding the specified threshold.
\end{itemize}

\begin{figure}[ht]
    \centering
    \includegraphics[width = \linewidth]{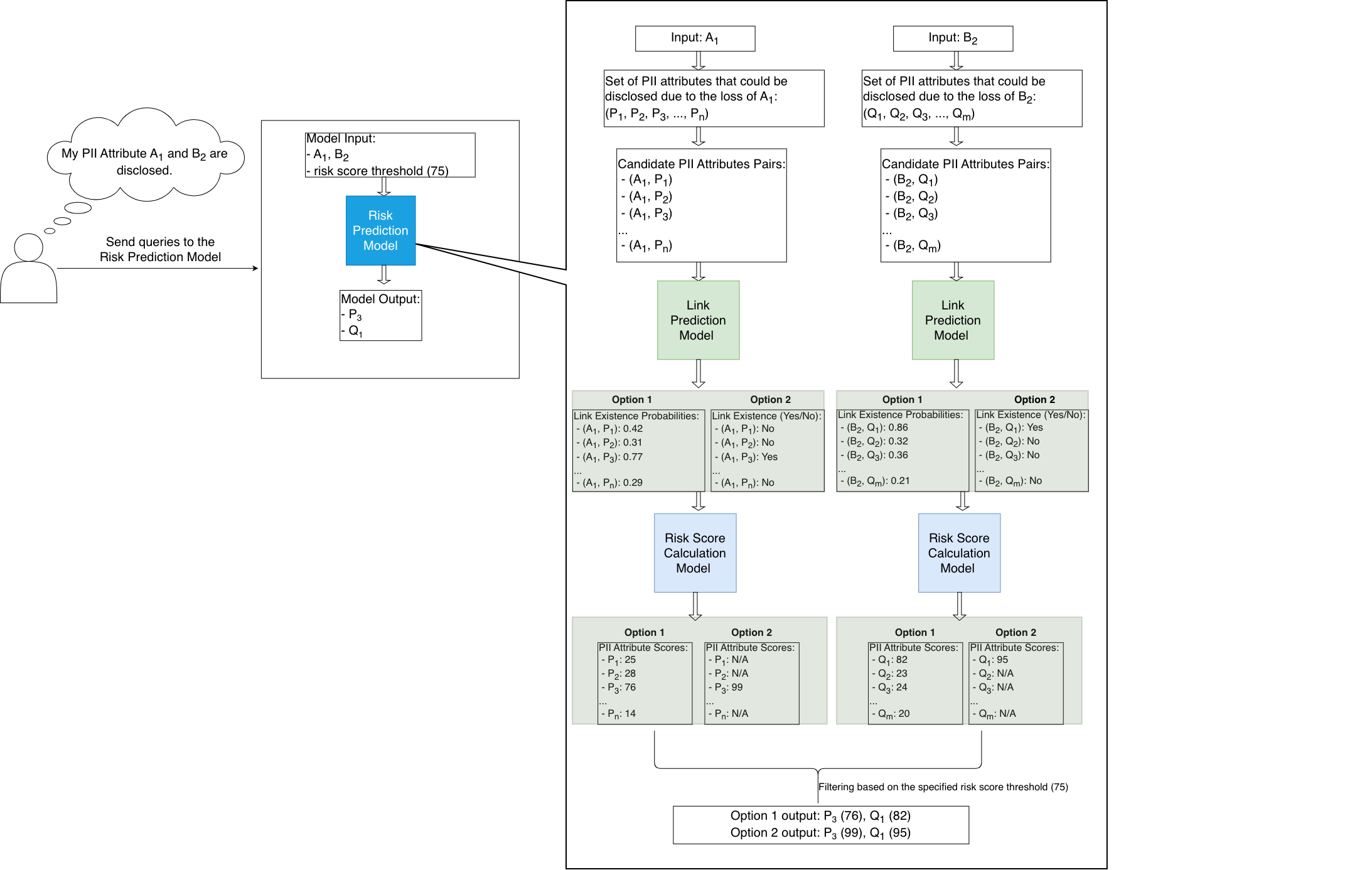}
    \caption{\textbf{Pipeline of the risk prediction framework}, illustrating the process from user query to PII attribute risk prediction results. User may choose different link prediction output formats depending on whether link prediction probabilities are heavily taken into account in the risk score calculation. Details of risk score calculation under different output options are provided in Section \ref{risk_calc_sec_6}.}
    \label{fig:pipeline}
\end{figure}

Figure \ref{fig:pipeline} uses an example of risk score threshold $75$ to explain the pipeline. In general, the contributions of this paper include:

\begin{itemize}
    \item We propose a method for constructing UTCID Identity Ecosystem graphs and provide mechanisms to customize the graphs for different scales and individual needs. (Section \ref{experimental_sec_3}).
    \item We create and train three different link prediction models (Section \ref{sec_4_2}, \ref{sec_4_3}, and \ref{sec_4_4}).
    \item We conduct extensive evaluations on UTCID Identity Ecosystem graphs of different sizes to demonstrate the performance of the proposed link prediction models. (Section \ref{result_sec_5}). 
    \item We construct a risk score calculation model to quantify privacy risks associated with PII disclosures (Section \ref{risk_calc_sec_6}).
\end{itemize} 

\section{\uppercase{Related Work}}
\label{related_work_sec_2}

Privacy protection and risk assessment are important research topics for many market sectors that collect, store, and analyze personally identifiable information. Researchers have developed a variety of data mining techniques to find suspicious patterns in data and identity fraud transactions \cite{javaid2024improving}. Applications of machine learning and data mining algorithms can help market sectors such as financial services, healthcare, transportation, and others mitigate financial losses and time costs. 

For instance, the paper \cite{zhang2024privacy} provides a method to assess privacy risks for medical big data. The researchers also apply the Fuzzy C-means clustering algorithm to cluster users into different groups, assign different permissions, and improve the access control accuracy. 

Despite substantial research on privacy risk assessment, there is limited work on identifying the connections among different aspects of personal data. This paper not only fills this gap by uncovering interconnections between personal data attributes, but also employs graph-based models to represent and predict their interactions, thereby revealing the risks associated with data sharing and leakage.

In relation to the graph-based models utilized in this project, link prediction methods play an important role. Link prediction is commonly used to detect missing links or add future connections. There are some simple methodologies based on node similarity, such as Jaccard’s coefficient \cite{murphy1996finley} and the Adamic/Adar measure \cite{adamic2003friends}. These techniques are computationally efficient but do not perform well across many types of graphs. Therefore, an increasing number of researchers have turned to machine learning algorithms \cite{wang2007local, lichtenwalter2010new, al2006link}.

The framework of this paper combines both simple similarity-based properties/scores and supervised learning algorithms, enabling efficient prediction of PII risks.

\section{\uppercase{UTCID Identity Ecosystem Graph Construction}}
\label{experimental_sec_3}

\begin{figure}[ht]
    \centering
    \includegraphics[width=\linewidth]{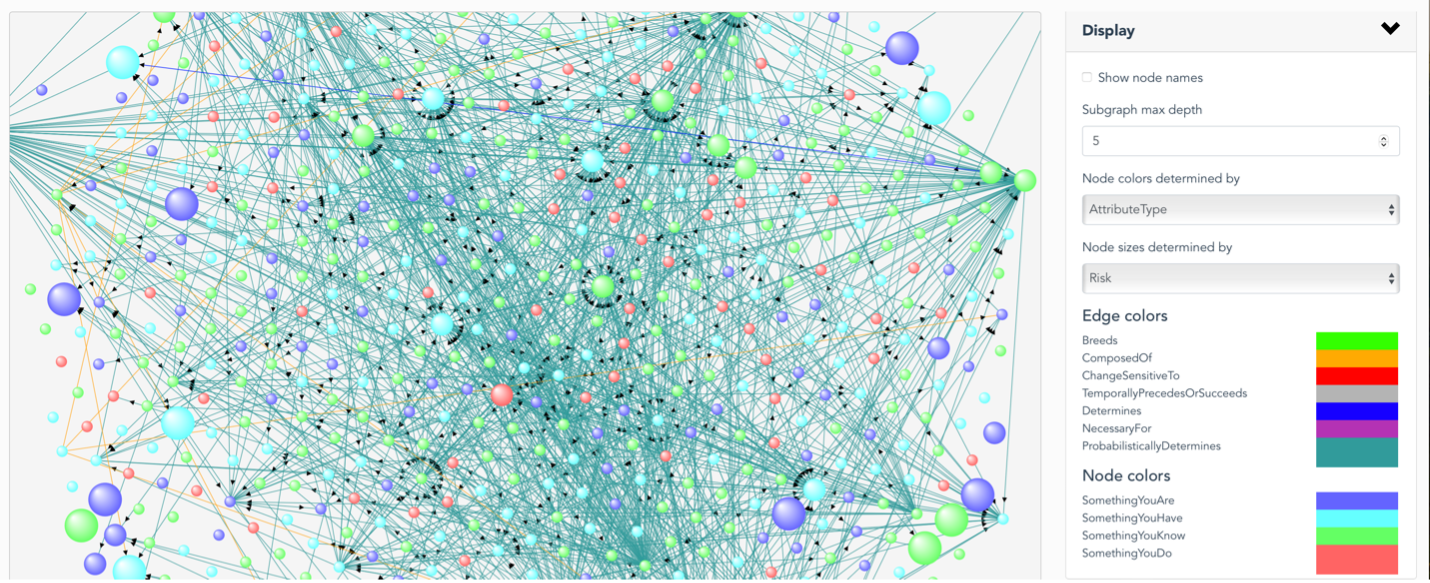}
    \caption{UTCID Identity Ecosystem Graph Representing PII Attributes and Their Relationships.}
    \label{fig:utcid-graph}
\end{figure}

We now explore the UTCID Identity Ecosystem---its content, its privacy risk analytics, and the new prediction capabilities introduced in this paper to help individuals protect their personal information from unintended and harmful disclosure. The Center for Identity at The University of Texas at Austin (UTCID) first introduced an Identity Ecosystem graph \cite{zaeem2016predicting}. The UTCID Identity Ecosystem graph represents PII attributes (i.e., personal data) as nodes and connects these nodes based on various types of relationships between PII attributes (Figure \ref{fig:utcid-graph}). While the work in \cite{zaeem2016predicting} presents multiple types of relationships, this paper focuses specifically on the \textit{probability of disclosure} relationship for privacy risk prediction. 

Figure \ref{fig:utcid-graph} illustrates the UTCID Identity Ecosystem graph, with nodes colored by Type (What You Know, What You Have, What You Are, What You Do) and sized by value. Based on more than $5,000$ identity theft and fraud cases collected in the Identity Threat Assessment and Prediction (ITAP) project \cite{zaeem2016predicting, zaiss2019identity} and analyzed to build the UTCID Identity Ecosystem graph, there is evidence that PII attributes have relative value. In an effort to estimate the relative contribution of a PII attribute $A_i$ to cumulative monetary loss across cases in which $A_i$ was used or targeted, the value of $A_i$ can be estimated as Equation \ref{monetization_loss_eq}. Let $S_{A_i}$ denote the set of cases in which $A_i$ appears, and $C_j$ $\in$ $S_{A_i}$ denotes an individual case.

\begin{equation}
    value(A_i) = \sum_{C_j\in S_{A_i}} \frac{MonetaryLoss(C_j)}{NumbrOfAttributes(C_j)}
\label{monetization_loss_eq}
\end{equation}

For simplicity, we refer to the collection of identity theft and fraud cases from the ITAP project as the ``UTCID ITAP dataset''. 

In this paper, we reconstruct the UTCID Identity Ecosystem graph and retain only a minimal set of structural graph features in order to predict PII disclosure risks based solely on PII attribute occurrences in the ITAP dataset. The construction of the new Identity Ecosystem graphs---referred to as UTCID Identity Ecosystem v2.0\footnote{UTCID Identity Ecosystem v2.0 offers a new means by which Identity Ecosystem graphs are constructed to promote accurate prediction of risks resulting from identity attribute exposure and disclosures. Throughout this paper, references to the UTCID Identity Ecosystem refer to UTCID Identity Ecosystem v2.0.}---follows the principles outlined below. Future work will examine how these prediction of PII risks can be used to predict harms such as monetary loss. 

We use nodes to represent PII attributes. A directed edge between PII attributes $A$ and $B$, denoted as $A \rightarrow B$, indicates that an event disclosing attribute $A$ leads to the disclosure of attribute $B$. Each edge is assigned a weight. Specifically, an edge $A \rightarrow B$ with weight $w_i$ indicates that an event disclosing $A$ may lead to the disclosure of $B$, and that such a disclosure occurred $w_i$ times in the empirical UTCID ITAP dataset \cite{zaiss2019identity}. 

\begin{figure}[ht]
    \centering
    \includegraphics[width=0.5\linewidth]{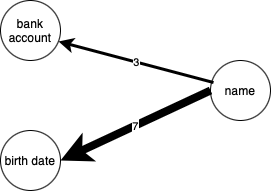}
    \caption{An Example Identity Ecosystem Graph with Three Nodes.}
    \label{fig:three-node}
\end{figure}

Figure \ref{fig:three-node} presents an example visualization based on a simplified use case, where edge thickness corresponds to the edge weight, representing the frequency of a specific disclosure relationship (i.e., $A \rightarrow B$ with weight $w_i$). This figure illustrates a simple UTCID Identity Ecosystem graph with three PII attributes. The weight of the edge ``name $\rightarrow$ bank account'' is $3$, indicating three observed events in which disclosure of a name led to the disclosure of a bank account. Similarly, the weight of the edge ``name $\rightarrow$ birth date'' is $7$, indicating seven observed events in which disclosure of a name led to the disclosure of a birth date. Assuming that these weights are derived from distinct events, the example graph illustrates the following:

\begin{itemize}
    \item Disclosure of the ``name'' attribute may lead to the disclosure of a ``bank account'' with probability $0.3$ (calculated as $3/(3+7) = 0.3$). 
    \item Disclosure of the ``name'' attribute may lead to the disclosure of a ``birth date'' with probability $0.7$ (calculated as $7/(3+7) = 0.7$).
\end{itemize}

Throughout this paper, we use UTCID Identity Ecosystem graphs, constructed according to the rules described above, to analyze and predict PII attribute disclosures and, consequently, privacy risks. 

To collect and structure the data necessary to perform the proposed risk prediction, two steps are required. First, we preprocess the UTCID ITAP dataset. For the inputs and outputs of the identity theft and fraud cases in the ITAP dataset, only identity-related information is retained. For each case, inputs are the data used by the actors to conduct the identity theft or fraud. The outputs include the consequent data that are acquired, stolen, or otherwise exposed by the conclusion of the identity theft or fraud actions. 

The second step is to construct the UTCID Identity Ecosystem graphs. We previously introduced the graph construction rules using a small example graph (e.g., Figure \ref{fig:three-node}, which illustrates a graph with three nodes and two edges). Here, we provide a more comprehensive explanation of these rules using multiple input and output identity attributes. 

Suppose there is a reported identity theft and fraud case. From this case, three input PII attributes can be extracted: ``bank account'', ``name'', and ``Social Security Number'' (abbreviated as ``SSN''). The corresponding output PII attributes are ``credit card'' and ``debit card''. If we construct a UTCID Identity Ecosystem graph based on this single case, five nodes and six directed edges are added to the graph. The five nodes are ``bank account'', ``name'', ``SSN'', ``credit card'', and ``debit card''. The six directed edges are listed below:
\begin{itemize}
    \item From ``bank account'' to ``credit card''.
    \item From ``bank account'' to ``debit card''.
    \item From ``name'' to ``credit card''.
    \item From ``name'' to ``debit card''.
    \item From ``SSN'' to ``credit card''.
    \item From ``SSN'' to ``debit card''.
\end{itemize}

Because this graph is constructed from a single identity theft and fraud case, every edge in the graph has a weight $1$, reflecting that each input--output attribute pair appears only once within this scope. 

Now consider a more complex scenario involving three identity theft and fraud cases (Cases 1, 2 and 3). In Case 1, the input PII attributes are ``bank account'', ``name'', and ``SSN'', and the corresponding output attributes are ``credit card'' and ``debit card''. In Case 2, the input attributes are ``bank account'' and ``SSN'', and the output attributes are ``birth date'', ``credit history'', and ``credit card''. In Case 3, the input attribute is ``SSN'', and the output attribute is ``bank account''. 

Across these three cases, there are seven unique PII attributes: ``bank account'', ``name'', ``SSN'', ``credit card'', ``debit card'', ``birth date'' and ``credit history''. Based on the three cases, the relationships among the seven attributes are listed below:
\begin{itemize}
    \item bank account $\rightarrow$ credit card (weight: $2$).
    \item bank account $\rightarrow$ debit card (weight: $1$).
    \item bank account $\rightarrow$ birth date (weight: $1$).
    \item bank account $\rightarrow$ credit history (weight: $1$).
    \item name $\rightarrow$ credit card (weight: $1$).
    \item name $\rightarrow$ debit card (weight: $1$).
    \item SSN $\rightarrow$ credit card (weight: $2$).
    \item SSN $\rightarrow$ debit card (weight: $1$).
    \item SSN $\rightarrow$ birth date (weight: $1$).
    \item SSN $\rightarrow$ credit history (weight: $1$).
    \item SSN $\rightarrow$ bank account (weight: $1$). 
\end{itemize}

\begin{figure}[ht]
    \centering
    \includegraphics[width=\linewidth]{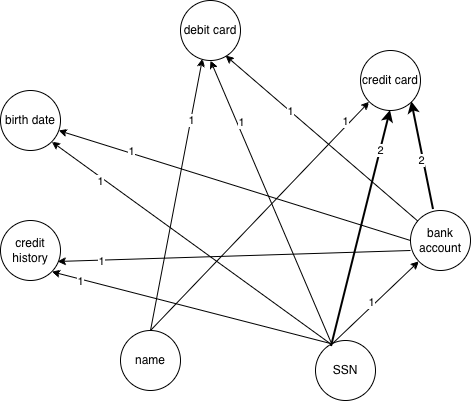}
    \caption{An Example of UTCID Identity Ecosystem Graph Construction Using Three Cases.}
    \label{fig:three-case}
\end{figure}

These weights reflect the occurrence frequencies of the input--output pairs among the identity theft and fraud cases. Figure \ref{fig:three-case} shows a visualization of the UTCID Identity Ecosystem graph constructed from the three-case example above. 

The largest UTCID Identity Ecosystem graph we construct in this paper is based on $5{,}636$ cases from the UTCID ITAP dataset. Following the procedures given above, the resulting graph contains $1{,}733$ nodes and $19{,}483$ edges. We denote this graph as $G_{grand}$.

Following the construction rules, UTCID Identity Ecosystem graphs can be constructed at different scales. For example, the UTCID ITAP dataset includes identity theft and fraud cases spanning a wide range of market sectors and victim demographics, involving different types of PII with varying values and losses. By filtering cases based on specific parameters (e.g., victim age, market sector, monetary loss amount), we can tailor the analysis to particular scenarios. As an illustration, filtering for cases with losses greater than $\$10{,}000$ yields a smaller graph with $761$ nodes and $6{,}413$ edges. We denote this filtered graph as $G_{big\_loss}$. 

\section{\uppercase{Link Prediction Algorithms}}
\label{algo_sec_4}

To answer a question such as \textbf{Could the disclosure of a given identity attribute possibly lead to the disclosure of another identity attribute of concern?}, we convert the question into a link prediction task \cite{kumar2020link} and a risk score calculation process, as shown in Figure \ref{fig:pipeline}. For the link prediction task, we aim to determine whether a directed edge exists between an initial PII attribute (input) and a target PII attribute (output). 

Suppose a UTCID Identity Ecosystem graph $G_{TX}$ is constructed based on identity theft and fraud cases that occurred in Texas. If an individual living in Texas is involved in a breach and is informed that their driver's license number was disclosed, then, given knowledge of $G_{TX}$, the link prediction algorithms can help determine whether a possible link (i.e., a potential privacy risk) exists between the disclosed ``driver's license number'' and other PII attributes, such as a bank account or a credit card. 

In general, link prediction algorithms can be used to examine whether directed links exist between initially exposed PII attributes and the remaining PII attributes in a UTCID Identity Ecosystem graph. An overview of link prediction algorithms is shown in Figure \ref{fig:summary} in the Appendix, constructed based on \cite{kumar2020link, arrar2024comprehensive}. 

Based on prior experimental studies conducted on ten benchmark homogeneous graphs (e.g., Ecoli \cite{salgado2001regulondb}, FB15K \cite{newman2006finding}), GNN-based link prediction algorithms have demonstrated superior performance compared with similarity-based methods \cite{islam2020comparative}. Additional research shows that graph convolutional networks (GCNs) and Word2Vec combined with a multilayer perceptron (MLP) offer significant computational efficiency advantages over exponential random graph--based approaches while maintaining strong prediction performance \cite{sosa2024unified}. 

Moreover, because the nodes of a UTCID Identity Ecosystem graph represent PII attributes and carry background information related to individuals’ identities, a major limitation of non-learning-based approaches is their difficulty in incorporating contextual information. Given the potentially large scale of UTCID Identity Ecosystem graphs and the importance of accurate privacy risk prediction, it is necessary to develop link prediction algorithms that are both efficient and accurate while accounting for graph structural complexity. Therefore, we narrow the link prediction algorithm choices to GNN-based and general deep-learning-based models, which are highlighted with red boxes in Figure \ref{fig:summary}. In the following subsections, we discuss the features of the UTCID Identity Ecosystem graphs that are used as inputs to the deep models. We also explain the three link prediction models that we propose.

\subsection{Semantic Processing for PII Attribute Nodes} \label{sec_4_1}

The nodes of UTCID Identity Ecosystem graphs correspond to PII attributes. These PII attributes are represented in natural language to describe identity- and privacy-related information. In other words, each PII attribute consists of one or more English words. In this paper, we focus exclusively on English-language PII attributes. 

Each English word has a corresponding definition that can be obtained from standard dictionaries. Accordingly, we define the \emph{contextual information} of PII attributes as the collection of word-level definitions associated with the words composing that attribute. We refer to the process of transforming a PII attribute from its surface form into a word-level semantic representation as \emph{semantic processing}. 

Our hypothesis is that, similar to inherent node properties of the UTCID Identity Ecosystem graphs (e.g., node degrees and node centrality), semantic features derived from PII attribute descriptions---such as token IDs obtained from word-level representations---provide meaningful node features. These semantic features can assist in predicting whether a link exists between two PII attribute nodes. We evaluate this hypothesis using the proposed model described in Section \ref{sec_4_4}. 

The semantic processing toolkit we use is the Natural Language Processing Toolkit (NLTK) \cite{bird2009natural}. For each node in a UTCID Identity Ecosystem graph, the semantic processing procedure consists of the following steps:
\begin{itemize}
    \item Extract the words that compose the PII attribute.
    \item Process each word iteratively using the NLTK \textit{synsets} and \textit{definition} functions to obtain a textual explanation. Specifically, we apply the \textit{synsets} function to obtain the synonym set for each word. We then concatenate the outputs of \textit{definition} corresponding to each item in the synonym set to derive the explanation of each word.  
    \item Concatenate the explanations of all constituent words into a single paragraph representing the contextual information of the PII attribute. 
\end{itemize}

Furthermore, we use the BERT-base-uncased tokenizer \cite{devlin2019bert} to obtain token IDs from the context information associated with each node. We treat the resulting token IDs as weak contextual signals---deterministic numerical representations that reflect semantic features and capture contextual patterns to a limited degree. Since our focus is not on comparing tokenization techniques or optimizing semantic embeddings, we do not evaluate alternative tokenizers in this paper. Our objective is to demonstrate that incorporating weak semantic information can improve link prediction performance. We refer to the token IDs generated from context information using the BERT-base-uncased tokenizer as \emph{semantic encoded signals} or \emph{semantic embeddings}, noting that this usage differs from the conventional definition of semantic embeddings.

\begin{figure*}[t]
    \centering
    \includegraphics[width=\textwidth]{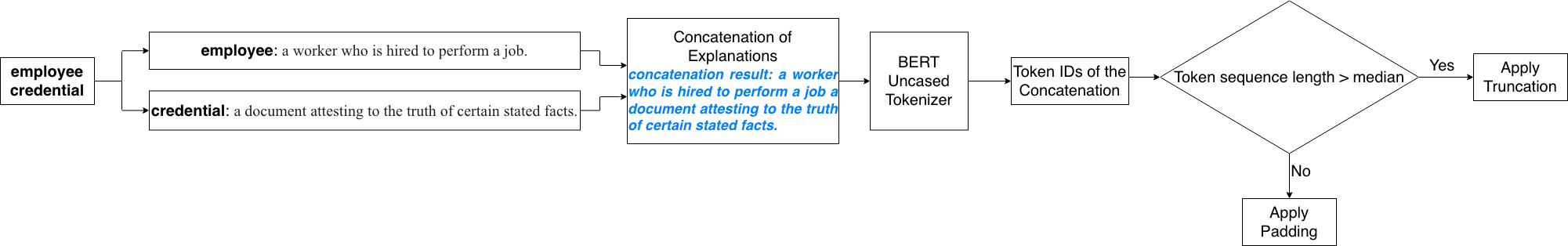}
    \caption{Process of Converting PII Attributes Expressed in English into Semantic Embeddings.}
    \label{fig:embedding}
\end{figure*}

To illustrate the process of converting PII attributes written in English into semantic embeddings, we consider a concrete example: the PII attribute ``employee credential'' (Figure \ref{fig:embedding}). This attribute consists of two words: ``employee'' and ``credential''. As described above, our semantic processing algorithm applies the NLTK \textit{synsets} and \textit{definition} functions to each word. Using this procedure, the context information for ``employee'' is ``a worker who is hired to perform a job'', and the context information for ``credential'' is ``a document attesting to the truth of certain stated facts''. Concatenating these two strings yields the context information for ``employee credential'': ``a worker who is hired to perform a job a document attesting to the truth of certain stated facts''. This text is then passed to the pretrained BERT-base-uncased tokenizer \cite{devlin2019bert} to generate the corresponding token ID sequence. The token ID sequences associated with different PII attribute nodes may vary in length. In our experiments, the median semantic embedding length across all nodes in $G_{grand}$ is $162$. Accordingly, we truncate sequences longer than $162$ and apply zero padding to sequences shorter than $162$. 

\subsection{MLP-Based Model for Link Prediction – FeatureMLP} \label{sec_4_2}

For each node in a directed graph, basic node properties such as in-degree and out-degree can be computed. These properties can be used to characterize structural similarity between nodes. Since links are more likely to exist between similar nodes \cite{al2022identifying}, and because powerful yet simple classifiers such as multilayer perceptrons (MLPs) \cite{rumelhart1986learning} and support vector machines (SVMs) \cite{cortes1995support} are effective in modeling nonlinear relationships, we construct an MLP-based link prediction model that uses basic node properties as features \cite{al2006link}. An additional motivation for using an MLP is to maintain architectural consistency among the deep learning models developed in this paper. We refer to this model as \emph{FeatureMLP}. 

\begin{figure}
    \centering
    \includegraphics[width=\linewidth]{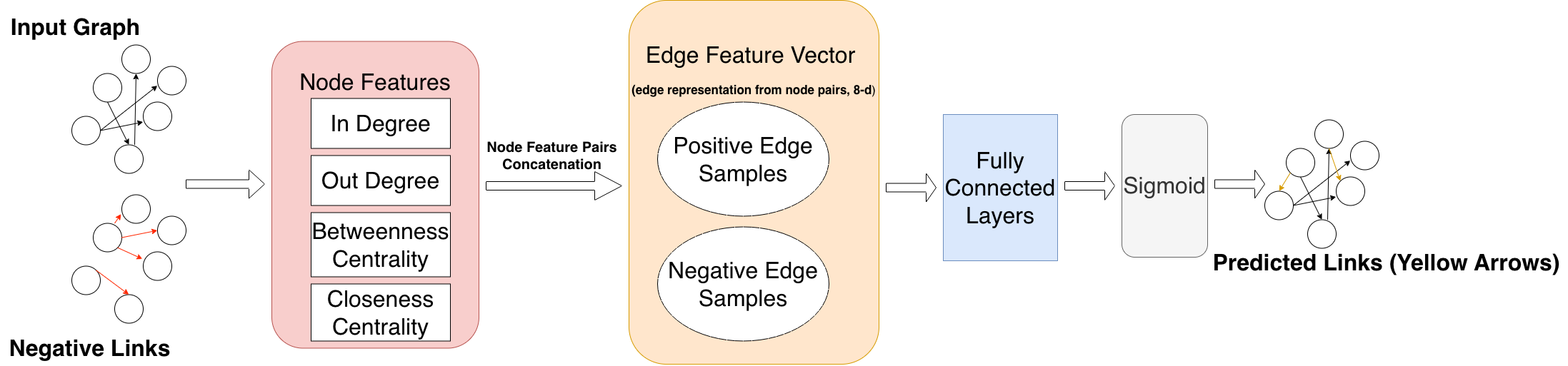}
    \caption{Overview of the FeatureMLP Model Architecture.}
    \label{fig:featureMLP}
\end{figure}

The overall model structure is shown in Figure \ref{fig:featureMLP}. For FeatureMLP, the node features (illustrated by the pink block in Figure \ref{fig:featureMLP}) include: 
\begin{itemize}
    \item \textbf{in-degree}: the number of incoming links for each node;
    \item \textbf{out-degree}: the number of outgoing links for each node;
    \item \textbf{betweenness centrality} \cite{freeman1977set}: a measure of how often a node lies on shortest paths between other nodes;
    \item \textbf{closeness centrality} \cite{sabidussi1966centrality}: a measure of how close a node is to all other nodes in the graph.
\end{itemize}

The edge feature vector is formed by concatenating the feature vector of a node pair. 

Since each node feature vector has dimension four, the resulting edge vector has dimension eight. 

FeatureMLP is computationally efficient and straightforward to implement, relying on low-level node statistics to perform link prediction. However, while these features capture global structural summaries, FeatureMLP does not explicitly model higher-order relational patterns or neighborhood-level interactions in the graph. To address scenarios that require capturing more complex structural dependencies and to maintain robust performance on large-scale graphs, we further develop additional link prediction models in Section \ref{sec_4_3} and \ref{sec_4_4}. 

\subsection{GCN-Based Model for Link Prediction – FeatureGCN} \label{sec_4_3}

Building on the first model (FeatureMLP), we next seek to leverage graph structural information. Therefore, we employ a two-layer graph convolutional network (GCN). Note that the term GCN is used here in a broad sense, encompassing successors of the original GCN architecture, including GraphSAGE \cite{hamilton2017inductive}.   

\begin{figure}
    \centering
    \includegraphics[width=\linewidth]{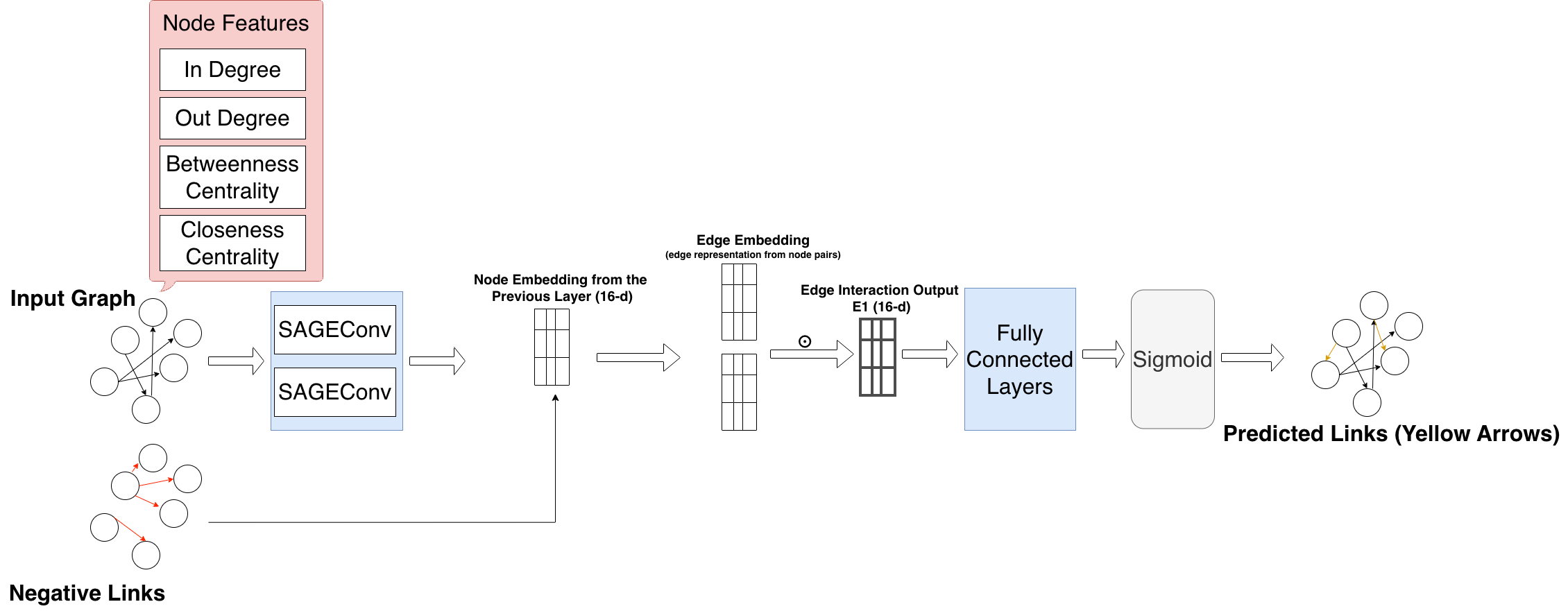}
    \caption{Overview of the FeatureGCN Model Architecture.}
    \label{fig:featureGCN}
\end{figure}

The model structure is shown in Figure \ref{fig:featureGCN}. We use two SAGEConv layers (based on GraphSAGE), implemented using the PyTorch Geometric package \cite{fey2019fast}, to generate node embeddings that capture local graph structural information. The output node embeddings have dimension $16$. We then combine the node embeddings to form edge embeddings. For each edge embedding, the corresponding node embeddings are denoted as $(ne_1, ne_2)$. We apply element-wise multiplication to incorporate the two node embeddings, resulting in a $16$-dimensional edge interaction output $ne_1 \odot ne_2$. This interaction output is denoted by $E1$ in Figure \ref{fig:featureGCN}. 

FeatureGCN simultaneously leverages low-level node properties and graph structural information. However, as discussed earlier, the semantic information associated with PII attributes also contains valuable signals. Therefore, a more advanced GCN-based model is required to incorporate semantic information, which we introduce in the next section. 

\subsection{GCN with Semantic Embeddings for Link Prediction - SeeGCN} \label{sec_4_4}

\begin{figure*}[ht]
    \centering
    \includegraphics[width=\linewidth]{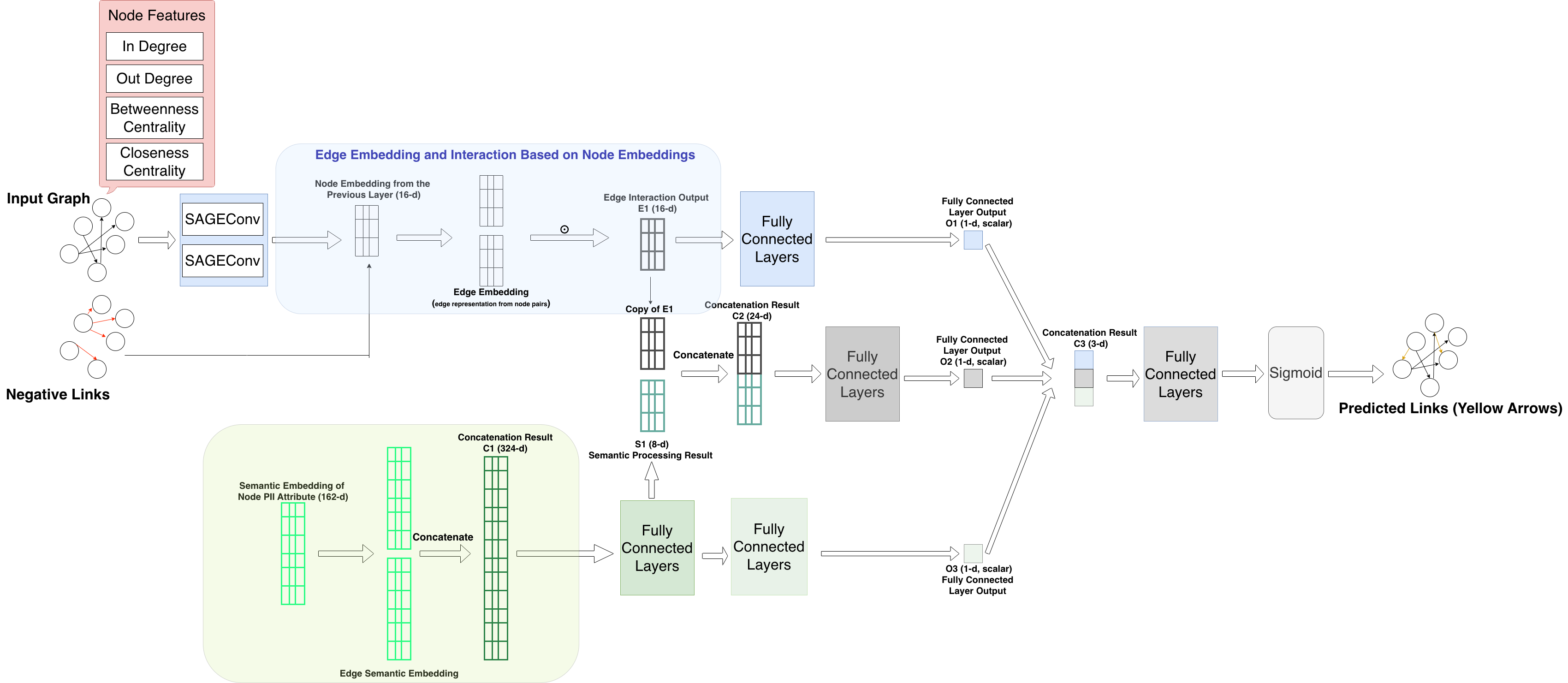}
    \caption{\textbf{Overview of the SeeGCN Model Architecture.} Normalizing the edge semantic embeddings before subsequent processing improves robustness across different graphs.}
    \label{fig:seeGCN}
\end{figure*}

As discussed in Section \ref{sec_4_1}, we obtain weak semantic embeddings---namely, the token ID sequences derived from the explanations of PII attributes---for each node in the input graphs. Building on the FeatureGCN model, we incorporate these semantic embeddings into the model and demonstrate that they can improve link prediction performance. The proposed model structure is shown in Figure \ref{fig:seeGCN}, and we refer to this new model as SeeGCN. 

For each pair of nodes $(n_1, n_2)$ corresponding to an edge in the input graphs, we obtain a pair of $162$-dimensional node semantic embeddings $(se_1, se_2)$. We refer to this pair as the edge semantic embedding. We then combine $se_1$ and $se_2$ through concatenation followed by fully connected layers. Note that normalizing the concatenated semantic token ID sequence pair $(se_1, se_2)$, as well as intermediate semantic representations produced by the fully connected layers, leads to more robust performance across different graphs. The resulting semantic concatenation is denoted as $C1$ in Figure \ref{fig:seeGCN}. Passing $C1$ through fully connected layers yields an eight-dimensional semantic processing result, denoted as $S1$. 

The edge interaction output $E1$ in Figure \ref{fig:seeGCN} is identical to that defined in Figure \ref{fig:featureGCN}. We concatenate $E1$ and $S1$ to obtain the $24$-dimensional representation $C2$, as shown in Figure \ref{fig:seeGCN}.

Overall, $E1$, $C2$, and $C1$ are each passed through fully connected layers, producing $O1$, $O2$, and $O3$, respectively. $O1$, $O2$, and $O3$ are scalars. We concatenate these three scalars to form a three-dimensional vector, which is then used to generate the final prediction results. 

This link prediction model effectively integrates low-level node properties, graph structural information, and semantic information. Moreover, the model shares architectural components with FeatureGCN and remains straightforward to implement. 

\section{\uppercase{Results and Analysis}}
\label{result_sec_5}

To compare the performance of the three proposed models, we evaluate them on the $G_{grand}$ and $G_{big\_loss}$ graphs, as described in Section \ref{experimental_sec_3}. We also randomly sample identity theft and fraud cases from the UTCID ITAP dataset and construct graphs based on these samples. We denote these graphs as $G_{(\#nodes,\#edges)}$, where $\#nodes$ represents the total number of nodes and $\#edges$ represents the total number of edges in the graph. To examine model robustness across graphs of different sizes, the number of randomly sampled cases ranges from $500$ to $5{,}000$, with evenly spaced intervals. Graphs are constructed using the NetworkX \cite{hagberg2008exploring} Python library and converted to PyTorch Geometric (PyG) data using the \textit{from\_networkX} function provided by PyG \cite{fey2019fast}. The number of training epochs is set to $100$, and the learning rate is set to $0.001$, which is the default value for the Adam optimizer. For each graph, edges are randomly split into training and validation sets using the \textit{RandomLinkSplit} function from the \textit{transforms} module of PyG, with a training-to-validation ratio of $9{:}1$. The model performance results, based on the experiments we have conducted, are reported in Table \ref{tab:result}. All experiments follow a consistent training and validation data construction strategy, as documented in the released code repository. 

\begin{table*}
    \centering
    \begin{tabular}{||c|c|c|c|c|c|c|c||}
    \hline
    \hline
       \multirow{2}{*}{\textbf{graph}}  & \multirow{2}{*}{\textbf{number of cases involved}} & \multicolumn{2}{|c|}{\textbf{FeatureMLP}} & \multicolumn{2}{|c|}{\textbf{FeatureGCN}} & \multicolumn{2}{|c||}{\textbf{SeeGCN}} \\
    \cline{3-8}
    & & AUC & ACC & AUC & ACC & AUC & ACC\\
    \hline
    \hline
    $G_{grand}$ & $5{,}636$ & $\textbf{0.94}$ & $0.84$ & $0.88$ & $0.83$ & $0.93$ & $\textbf{0.85}$ \\
    \hline
    $G_{big\_loss}$ & $897$ & $\textbf{0.92}$ & $0.80$ & $0.86$ & $0.82$ & $0.90$ & $\textbf{0.83}$ \\
    \hline
    $G_{(611,3094)}$ & $500$ & $\textbf{0.95}$ & $\textbf{0.87}$ & $0.89$ & $0.84$ & $0.91$ & $0.83$ \\
    \hline
    $G_{(823,5135)}$ & $1{,}000$ & $\textbf{0.94}$ & $\textbf{0.87}$ & $0.91$ & $0.84$ & $0.92$ & $0.85$ \\
    \hline
    $G_{(1003,7221)}$ & $1{,}500$ & $\textbf{0.93}$ & $\textbf{0.85}$ & $0.90$ & $0.77$ & $0.92$ & $0.83$ \\
    \hline
    $G_{(1128,9095)}$ & $2{,}000$ & $\textbf{0.93}$ & $\textcolor{red}{0.65}$ & $0.91$ & $\textbf{0.85}$ & $\textbf{0.93}$ & $\textbf{0.85}$ \\
    \hline
    $G_{(1213,10647)}$ & $2{,}500$ & $\textbf{0.94}$ & $\textbf{0.83}$ & $0.91$ & $\textbf{0.83}$ & $0.90$ & $\textbf{0.83}$ \\
    \hline
    $G_{(1318,12279)}$ & $3{,}000$ & $\textbf{0.94}$ & $\textcolor{red}{0.61}$ & $0.89$ & $0.82$ & $0.92$ & $\textbf{0.84}$ \\
    \hline
    $G_{(1398,13689)}$ & $3{,}500$ & $\textbf{0.94}$ & $\textcolor{red}{0.53}$ & $0.89$ & $0.84$ & $0.91$ & $\textbf{0.85}$ \\
    \hline
    $G_{(1471,15018)}$ & $4{,}000$ & $\textbf{0.94}$ & $0.77$ & $0.86$ & $0.82$ & $0.93$ & $\textbf{0.86}$ \\
    \hline
    $G_{(1554,16413)}$ & $4{,}500$ & $\textbf{0.93}$ & $\textcolor{red}{0.61}$ & $0.90$ & $\textbf{0.85}$ & $\textbf{0.93}$ & $\textbf{0.85}$ \\
    \hline
    $G_{(1625,17694)}$ & $5{,}000$ & $\textbf{0.94}$ & $\textbf{0.86}$ & $\textcolor{red}{0.68}$ & $\textcolor{red}{0.64}$ & $0.92$ & $\textbf{0.86}$ \\
    \hline
    \end{tabular}
    \caption{Model Evaluation Results on Different Graphs; AUC---ROC AUC score, ACC---Accuracy.}
    \label{tab:result}
\end{table*}

The table summarizes the performance of different models across the evaluated graphs. For each experiment, we report the best validation performance achieved over $100$ training epochs in terms of ROC AUC \cite{hanley1982meaning} and link existence prediction accuracy. For each graph, the highest AUC and accuracy values are highlighted in bold. Values below $0.7$ are highlighted in red to indicate relatively poor results. 

Across all experiments, at least two of the three proposed models achieve strong predictive performance, with AUC scores above $0.8$. Although we designate scores below $0.7$ as relatively poor, all reported performances remain above the random-guessing baseline of $0.5$. These results indicate that all three proposed models demonstrate robust predictive capability. Notably, the SeeGCN model does not exhibit any results below the $0.7$ threshold, suggesting strong robustness across both AUC and accuracy metrics. Based on the results reported in Table \ref{tab:result}, FeatureGCN and SeeGCN generally maintain stable link prediction performance on larger graphs, with the exception that FeatureGCN performs poorly on the graph constructed from $5{,}000$ cases ($G_{(1625,17694)}$). Overall, the findings show that link relationships in UTCID Identity Ecosystem graphs can be effectively predicted using the proposed models. 

In this paper, UTCID Identity Ecosystem graphs are constructed from empirical cases in the UTCID ITAP dataset. The original dataset contains manual annotation errors and inconsistencies. While some of the errors are removed during preprocessing, the dataset is intentionally not heavily cleaned, resulting in graphs that contain noise. Despite this, the models achieve AUC scores as high as $0.95$, and at least two of the three proposed models attain AUC values above $0.80$ in every experiment. These results suggest that the proposed link prediction framework is robust to data noise. Moreover, they indicate that prediction performance could further improve when using cleaner data or when users construct customized UTCID Identity Ecosystem graphs from higher-quality empirical cases.

\section{\uppercase{Risk Score Calculation}}
\label{risk_calc_sec_6}

For the same PII attribute, people may assign different levels of importance. Some people may spend most of their time on social media, while others may only check social media occasionally. It is therefore reasonable that individuals in the former group would assign higher risk scores to usernames and passwords associated with their online accounts. After identifying potentially disclosed nodes based on the link prediction results from Section \ref{algo_sec_4}, risk scores can be assigned either manually according to individual preferences or automatically using quantitative evaluation metrics. 

In this section, we propose a risk score calculation method. Suppose the query is to assess the risk scores of nodes related to the disclosure of a PII attribute $\alpha$. Let the PII attribute set be defined as all nodes in a UTCID Identity Ecosystem graph except for $\alpha$, denoted as $\{n_1, n_2, n_3, \ldots, n_m\}$, where $m$ is the total number of attributes in the set. 

We apply the PageRank algorithm \cite{page1999pagerank} to the graph of interest to obtain PageRank coefficients, denoted as $PR(n_i)$ for each node $n_i$. We also calculate the reverse PageRank coefficients by applying the reverse PageRank algorithm on the graph \cite{bar2008local}, denoted as $rPR(n_i)$ for each node $n_i$. The score $S_i$ for each node is defined as the sum of its forward and reverse PageRank coefficients: 

\begin{equation}
    S_i = PR(n_i) + rPR(n_i).
\end{equation}

Let $maxS$ denote the maximum value of $S_i$ across all nodes. We normalize and scale $S_i$ to the range $[0, 100]$ as follows:

\begin{equation}
    S_i= \frac{S_i}{maxS}\times100.
    \label{scale_result_1}
\end{equation}

Next, for each node pair $(\alpha, n_i)$, the link prediction algorithm outputs either a link existence probability $p_i$ or a binary indicator of link existence. If a user chooses to output the link existence probability, the risk score of node $n_i$ under the disclosure of $\alpha$ is defined as:

\begin{equation}
    RS_i= p_i\times S_i.
\end{equation}

Let $maxRS$ denote the maximum value of $RS_i$ across all nodes. To map the risk scores to the range $[0, 100]$ scale, we normalize and scale $RS_i$ as: 

\begin{equation}
   RS_i= \frac{RS_i}{maxRS}\times100. 
   \label{scale_result_2}
\end{equation}

\begin{figure*}
    \centering
    \includegraphics[width=\linewidth]{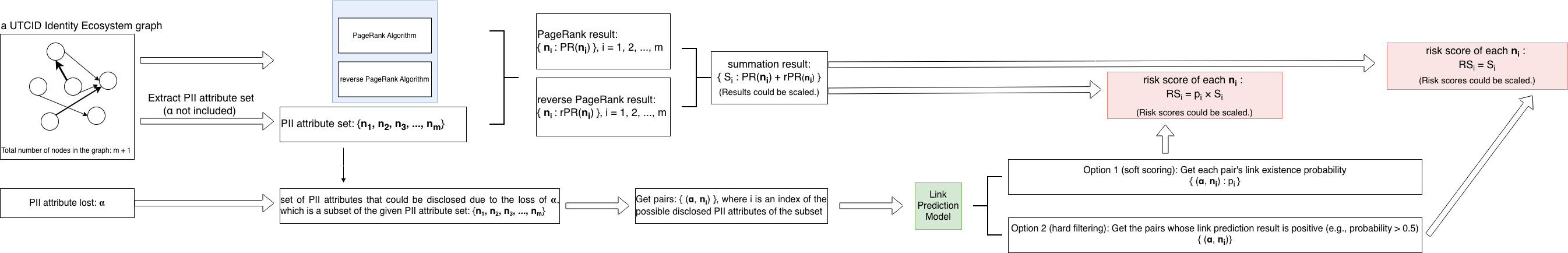}
    \caption{Process of Risk Score Calculation.}
    \label{fig:risk_score}
\end{figure*}

Alternatively, if the user chooses to consider only whether a link exists, the risk score of node $n_i$ is set equal to $S_i$ when the link prediction result indicates a positive link from $\alpha$ to $n_i$. Figure \ref{fig:risk_score} illustrates the risk score calculation process under the two link prediction output options. 

We do not evaluate which link prediction output option is superior, as the two options are intended to provide flexibility to accommodate different user preferences. The difference between the two approaches is illustrated as follows. Suppose the link prediction probability for $(\alpha, n_{10})$ is $0.42$, and the scaled PageRank-based score for $n_{10}$ is $S_{10} = 97$. Assume the user wishes to protect the PII attributes with risk scores above $40$. 

When the link prediction probability is incorporated, the unscaled risk score is $97\times0.42 = 40.74$. Applying Equation \ref{scale_result_2} yields a final risk score of $81.48$, assuming $maxRS = 50$ in this example. In this case, the user should allocate resources (e.g., money, time, or effort) to protect the PII attribute corresponding to $n_{10}$. In contrast, if the user considers only positive link prediction outcomes---using a probability threshold of $0.5$ as an example in this paper---then $n_{10}$ would not be selected for protection, since its link prediction probability is $0.42$. 

Although we provide the equations to convert the scores to a $[0, 100]$ scale (Equations \ref{scale_result_1} and \ref{scale_result_2}), users are free to adopt alternative scales to reflect their own perceptions of PII attributes. 

\section{\uppercase{Conclusion}}
\label{conclusion_sec_7}

The goal of this paper is to analyze and predict privacy risks incurred when individuals share personal data. We introduce MLP-based and GCN-based algorithms---FeatureMLP, FeatureGCN, and SeeGCN---to answer the following question: \textbf{Can the disclosure of a given identity attribute possibly lead to the disclosure of another attribute}? 

This work is motivated by the premise that individuals who better understand the risks associated with sharing specific personal data (identity attributes) are better equipped to make informed decisions and protect their privacy. Experimental evaluations on multiple UTCID Identity Ecosystem graphs demonstrate that the proposed framework can answer this question with strong predictive performance, robustness, and consistency. Furthermore, the UTCID Identity Ecosystem graphs, together with the proposed privacy risk prediction framework, offer flexibility and customization to support a wide range of practical use cases.

Future work will focus on identifying optimal GCN architectures for graphs of varying sizes, exploring integration between graph neural networks and reinforcement learning, and developing more advanced methods for incorporating semantic information from PII attributes into GCN-based link prediction models. 

\section{\uppercase{Acknowledgments}}
The authors used ChatGPT and Google Search AI tools solely for grammar check, language polishing, and manuscript review. All research ideas, methods, analyses, and conclusions are entirely those of the authors. 

\bibliographystyle{apalike}
\begingroup
\small
\bibliography{ref}
\endgroup

\onecolumn
\section*{\uppercase{Appendix - Link Prediction Methods Overview}}
\label{appendix: link prediction overview}

\begin{figure*}[ht]
    \centering
    \includegraphics[width = .8\linewidth]{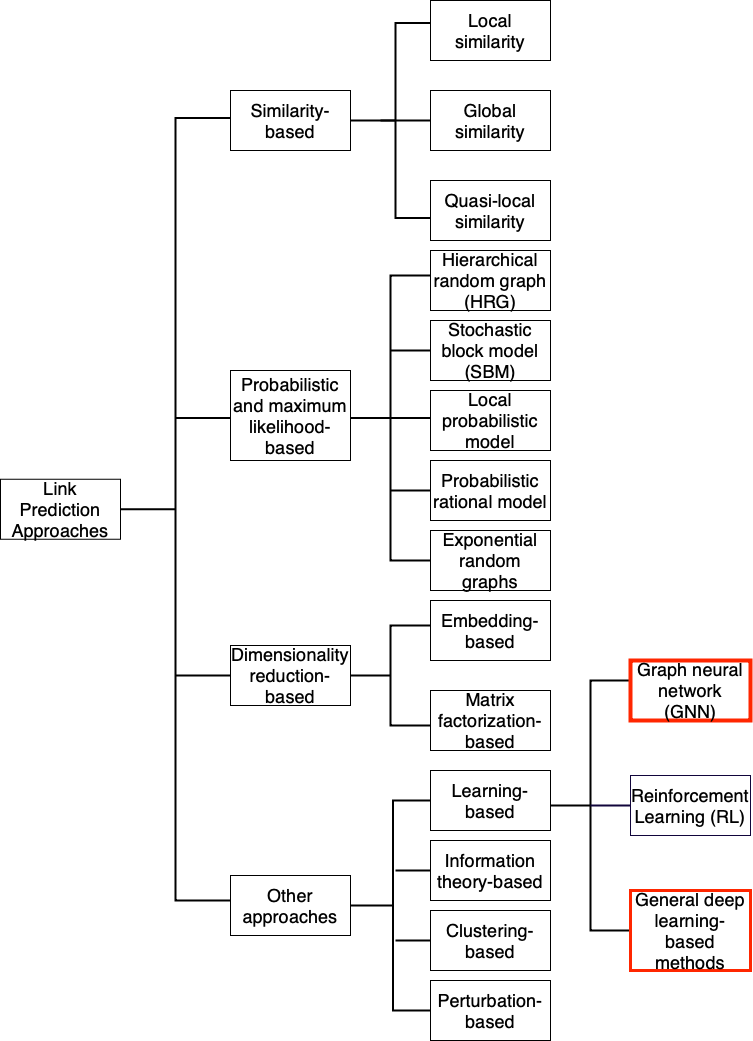}
    \caption{\textbf{Overview of Link Prediction Algorithms} \cite{kumar2020link, arrar2024comprehensive}. Reinforcement learning is shown as a representative learning-based algorithm in this discussion. }
    \label{fig:summary}
\end{figure*}

Figure \ref{fig:summary} shows an overview of link prediction algorithms, as discussed in Section \ref{algo_sec_4}. The red boxes highlight the focus of this paper. 

\end{document}